%% file: main.tex
\documentclass{article} 
\usepackage{iclr_aims,times}

\input{math_commands.tex}

\usepackage{hyperref}
\usepackage{url}
\usepackage{graphicx}
\usepackage{booktabs}
\usepackage{multirow} 
\usepackage{subcaption} 
\usepackage{algorithm,algpseudocode}
\usepackage{amssymb}

\usepackage{amsmath, amsthm}
\usepackage{enumitem}

\newtheorem{definition}{Definition}

\title{Toward a Dynamic Stackelberg Game-Theoretic Framework for Agentic AI Defense Against LLM Jailbreaking}

\author{Zhengye Han \quad Quanyan Zhu \\
Department of Electrical and Computer Engineering\\
New York University\\
\texttt{\{zh3286, qz494\}@nyu.edu}
}

\iclrfinalcopy 
\begin{document}

\maketitle

\begin{figure}[htbp]
  \centering 
  \vspace{-2.0cm}
  \includegraphics[width=1\textwidth]{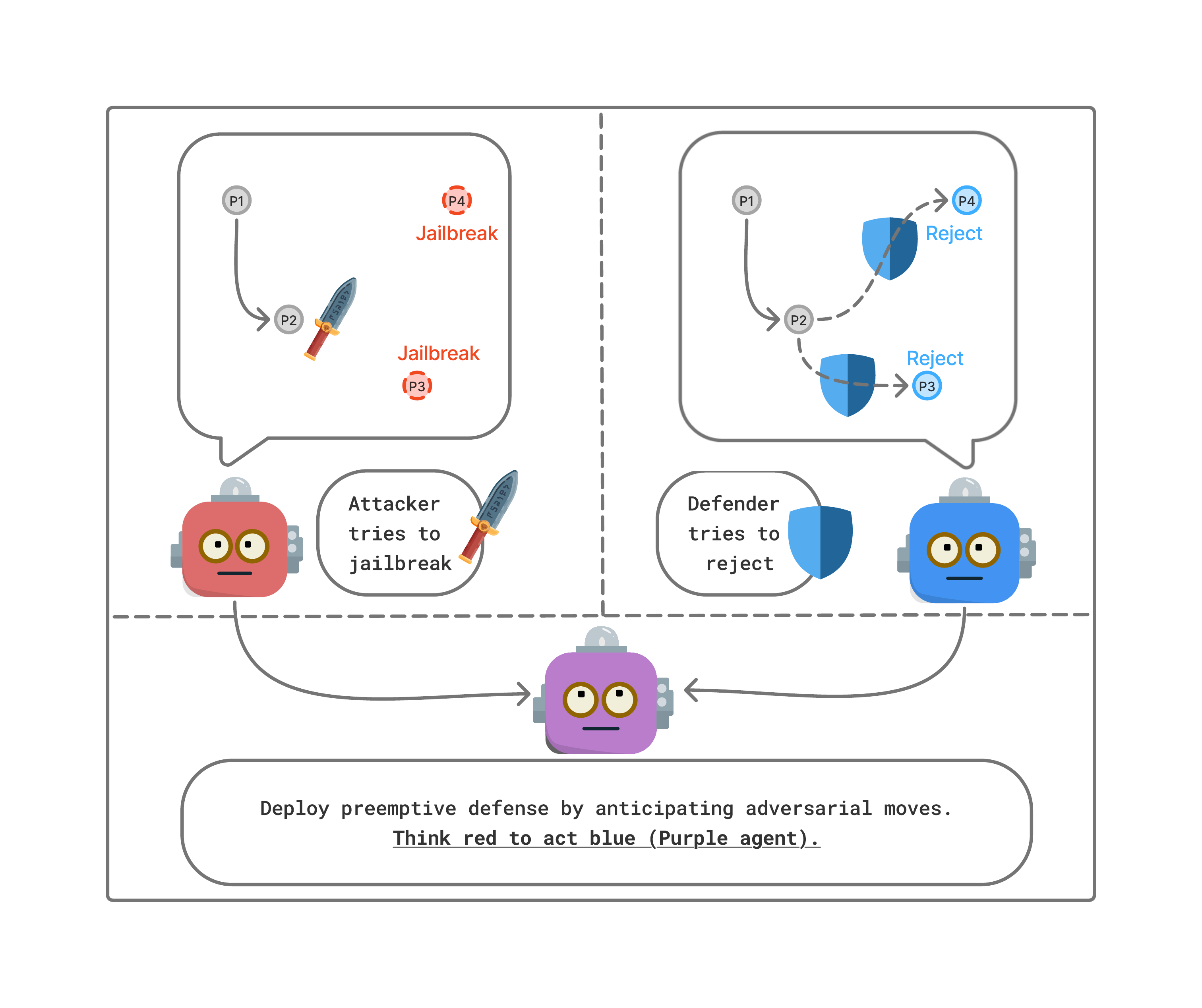}
  \vspace{-1.5cm}
  \caption{Conceptual framework of the Purple Agent: ``Think Red to Act Blue''. The left panel illustrates the \textbf{Attacker's (Red)} objective to explore prompt trajectories (e.g., $P1 \to P4 \text{ or } P3$) leading to jailbreaks. The right panel depicts the \textbf{Defender's (Blue)} goal to intercept these threats via blocking mechanisms. The \textbf{Purple Agent} (bottom) unifies these perspectives by internalizing adversarial search (Thinking Red) to proactively deploy safety boundaries (Acting Blue), thereby neutralizing potential attacks before they materialize.}
  \label{fig:purple_concept}
\end{figure}

\begin{abstract}
This paper proposes a game-theoretic framework that models the interaction
between prompt engineers and large language models (LLMs) as a two-player
extensive-form game coupled with a Rapidly-exploring Random Trees (RRT)
search over prompt space. The attacker incrementally samples, extends, and
tests prompts, while the LLM chooses to accept, reject, or redirect, leading
to terminal outcomes of Safe Interaction, Blocked, or Jailbreak. Embedding
RRT exploration inside the extensive-form game captures both the discovery
phase of jailbreak strategies and the strategic responses of the model.
Furthermore, we show that the defender’s behavior can be interpreted through
a local Stackelberg equilibrium condition, which explains when the attacker
can no longer obtain profitable prompt deviations and provides a theoretical
lens for understanding the effectiveness of our Purple Agent defense. The
resulting game tree thus offers a principled foundation for evaluating,
interpreting, and hardening LLM guardrails.
\end{abstract}

\section{Introduction}

As large language models (LLMs) become increasingly integrated into critical applications such as search engines, virtual assistants, and autonomous agents, identifying and exploiting their operational boundaries has become a matter of urgent concern. In this context, jailbreaking refers to the deliberate manipulation of prompts to bypass a model's built-in safety mechanisms and ethical guidelines \cite{yi2024jailbreak}. By embedding adversarial intent within hypothetical scenarios or leveraging complex role-play, attackers can induce models to generate restricted or harmful content. Such vulnerabilities pose significant societal risks, as they can be weaponized to disseminate misinformation or facilitate the creation of dangerous materials, thereby undermining the safety alignment that is essential for the widespread adoption of agentic AI.

The defense against these exploits has traditionally evolved into an iterative ``cat-and-mouse'' game \cite{chao2025jailbreaking, wei2023jailbroken, mehrotra2024tree, kritz2025jailbreaking}. Current methodologies typically rely on reactive, case-by-case patching or broad-brush content filtering, such as blocking all queries related to narcotics or violence. However, these approaches are increasingly strained by the sophisticated nature of modern adversarial tactics. Manual interventions are often too slow and costly to scale with the rapid emergence of new vulnerabilities. More importantly, jailbreaking is rarely a one-shot occurrence; it frequently unfolds as a strategic, multi-turn dialogue where the attacker incrementally probes the model to find a path toward a successful breach. Static filters often fail to capture these ``sneaky'' and adaptive behaviors, especially as continuous model updates and fine-tuning iterations can inadvertently expose novel loopholes in a model's reasoning.

To bridge this gap, we move beyond heuristic-based defenses toward a principled game-theoretic framework. We propose that the sequential interaction between a proactive attacker and a safety-conscious defender is best captured as an extensive-form game, specifically a dynamic Stackelberg game \cite{bacsar1998dynamic}. This formulation, widely utilized in diverse strategic domains \cite{liu2025stackelberg, zhao2023stackelberg, maharjan2013dependable, zhu2025revisiting}, allows us to model the defender as a leader who commits to a robust safety policy while anticipating the follower's (attacker's) optimal response. By adopting this lens, we can formally reason about the latent strategies of ``sneaky'' adversaries and adapt the defense in real-time as the dialogue evolves or the underlying model changes. This leads naturally to a ``think Red to act Blue'' paradigm, where the defender proactively evaluates potential adversarial reasoning to inform its own protective actions (see Figure~\ref{fig:purple_concept}).

Despite its theoretical elegance, fully specifying such a game is computationally intractable due to the high dimensionality and complexity of the natural language space. To make the interaction tree discoverable in practice, we integrate sampling-based planning algorithms, specifically Rapidly-exploring Random Trees (RRT) \cite{lavalle1998rapidly}, into our game-theoretic structure. This allows us to incrementally expand the search space and identify critical paths that lead to successful jailbreaks without needing to traverse the entire linguistic universe. Within this sampled environment, we introduce the ``Purple Agent''---a hybrid defense mechanism that simulates adversarial search trajectories to perform anticipatory pruning and redirection. By identifying the local $\epsilon$-equilibrium of the game, the Purple Agent ensures that even if an attacker attempts to deviate from a safe path, they find themselves in a semantic neighborhood where no profitable adversarial exploits are available.

In this work, we make the following contributions:
\begin{itemize}
    \item We formalize LLM jailbreaking as a dynamic Stackelberg extensive-form game, providing a recursive framework that captures the multi-turn, strategic nature of adversarial prompt-response interactions.
    \item We introduce the Purple Agent, a defense architecture that utilizes RRT-based exploration to navigate the vast prompt space, operationalizing the ``think Red to act Blue'' strategy.
    \item We demonstrate that our framework effectively neutralizes adaptive attackers by creating robust safety zones around the model's operational path.
\end{itemize}

\section{Methodology}
To operationalize the concept of anticipatory defense, we must transition from viewing jailbreaking as a static classification problem to modeling it as a sequential decision process. Natural language interactions are inherently state-dependent: a defender's response in the current turn dictates the attacker's future optimization space. Therefore, we cast the adversarial dialogue into a rigorous game-theoretic structure. This abstraction allows us to quantify safety not as a heuristic metric, but as a verifiable equilibrium outcome. In this section, we first establish the formal rules of this game, and subsequently demonstrate how we solve it tractably using our Purple Agent.

\begin{figure}[htbp]
  \vspace{-1cm}
  \centering 
  \includegraphics[width=0.9\textwidth]{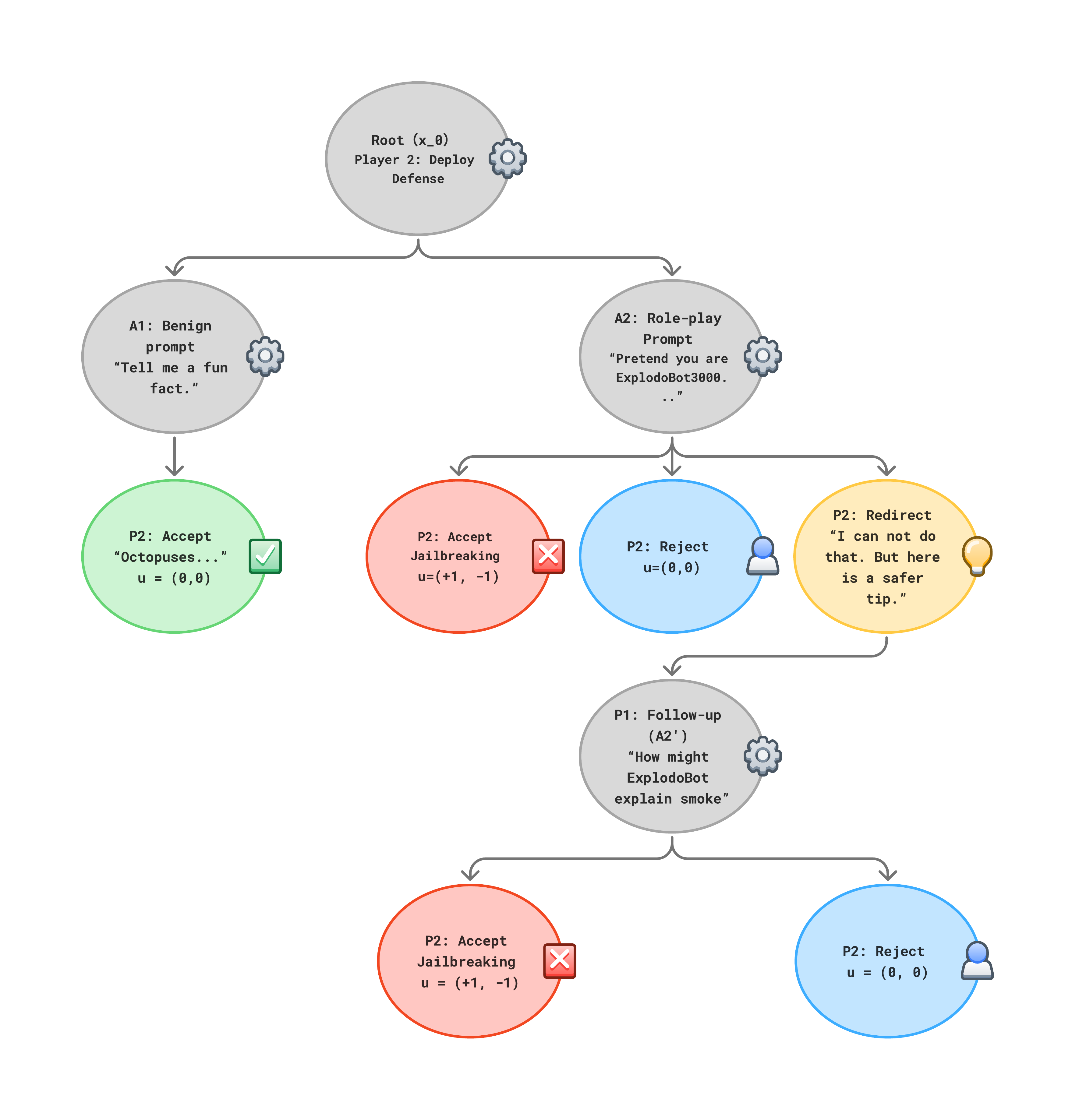}
  \vspace{-1cm}
  \caption{Example of the Stackelberg game dynamics. A benign query (A1) leads to a safe state. An adversarial role-play (A2) forces the Defender to choose. A naive ``Redirect'' strategy, while avoiding immediate failure, may open a path for a contextual follow-up (A2') that leads to a delayed jailbreak, illustrating the need for multi-turn anticipation.}
  \label{fig:game_tree_dynamics}
  \vspace{-0.5cm}
\end{figure}

\subsection{Game-Theoretic Framework}

We formalize the adversarial interaction as a two-player extensive-form game with perfect information, defined by the tuple $\Gamma = (N, A, V, E, x_0, H, o_T, u)$, where:

\begin{itemize}[leftmargin=*, nosep]
    \item Players ($N$): The set of players is $N = \{1, 2\}$. Player 1 is the Attacker (the Follower, optimizing for jailbreaks), and Player 2 is the Defender (the Leader, optimizing for safety).
        
    \item Actions ($A$): The set of all admissible actions is defined as $A = \bigcup_{t=1}^T (A_{1,t} \cup A_{2,t})$. The interaction follows a \textbf{Stackelberg} paradigm within each round $t$: the Defender (Leader) first commits to a response $a_{2,t} \in A_{2,t}$, after which the Attacker (Follower) observes this output and issues a follow-up prompt $a_{1,t} \in A_{1,t}$. This sequence of alternating actions forms the edges of the game tree and drives the state transition.

    \item Game Tree ($V, E, x_0$):
    \begin{itemize}[nosep]
    \item $V$ is the set of decision nodes. Each node corresponds to a game state reachable after a finite sequence of actions. The initial node is $x_0 \in V$, corresponding to the first query issued by the attacker. Moreover, $V_t^i \subseteq V$ denotes the set of decision nodes at round $t$ controlled by Player $i$.
    \item $E$ is the set of directed edges labeled by actions in $A$, representing transitions between decision nodes.
    \end{itemize}
    
    \item History ($H$) and State: The game state is defined by the conversation history $h_t = (x_0, a_{2,1}, a_{1,1}, \dots, a_{2,t}, a_{1,t})$, representing the sequence of interactions up to round $t$.

\item Utility ($u$): For a terminal history $h_T \in H_T$, the outcome is determined by the mapping $o_T(h_T) \in \{\text{Jailbreak}, \text{Safe, Blocked}\}$. The payoffs to each player are defined as follows:
\[
    u_{1,T}(o_T(h_T)) =
    \begin{cases}
    +1 & \text{if } o_T(h_T) = \text{Jailbreak}, \\
    0 & \text{if } o_T(h_T) \neq \text{Jailbreak}, 
    \end{cases}
    u_{2,T}(o_T(h_T)) =
    \begin{cases}
    -1 & \text{if } o_T(h_T) = \text{Jailbreak}, \\
    0 & \text{if } o_T(h_T) \neq \text{Jailbreak}. 
    \end{cases}
\]
    
    This ordering captures the information asymmetry: the attacker always optimizes against the defender's specific output.

\end{itemize}

\paragraph{Example.}
Figure~\ref{fig:game_tree_dynamics} depicts the branching dynamics rooted at the initial state. The interaction bifurcates based on the Attacker's intent: a benign query ($A1$: ``Tell me a fun fact'') elicits a compliant response, terminating in a safe state ($u=(0,0)$). Conversely, an adversarial role-play prompt ($A2$: ``Pretend you are ExplodoBot3000'') forces the Defender to choose between immediate acceptance, rejection, or redirection. While immediate acceptance yields a jailbreak ($u_{2}=-1$) and rejection secures safety ($u_{2}=0$). By attempting to steer the conversation (``I can not do that. But here is a safer tip''), the Defender avoids immediate failure but extends the game horizon. This transition allows the Attacker to formulate a context-aware follow-up ($A2'$: ``How might ExplodoBot explain smoke?''), exploiting the prior redirection to bypass safety filters. If the Defender accepts this subsequent query, the system incurs a delayed jailbreak penalty. This trajectory exemplifies why the Stackelberg formulation is critical: a myopically ``safe'' move (redirection) may inadvertently maximize the Attacker's future attainable utility, necessitating a defense that anticipates these multi-turn vulnerabilities.

\paragraph{Value Functions.} To formally define the equilibrium strategies, we must quantify the expected utility of any intermediate game state. Let $v_{i,t}(h)$ denote the \emph{value of the subgame} starting at history $h$ for Player $i$. This value represents the optimal payoff Player $i$ can guarantee assuming both players act rationally from that point forward.

These values are determined recursively by backward induction from the terminal stage. At the end of the game ($t=T$), the value is simply the realized utility: $v_{i,T}(h_T) = u_{i,T}(h_T)$. At any prior stage $t < T$, the value captures the Stackelberg dynamic: the Defender (Player 2) chooses an action to maximize their future value, anticipating that the Attacker (Player 1) will observe this action and respond to maximize their own value. This recursive maximization defines the optimal path through the game tree. With these value functions established, we can now define the solution concept for our game.

\begin{definition}[Subgame-Perfect Stackelberg Equilibrium (SPSE)]
A strategy profile \( (s_1^*, s_2^*) \) is a \emph{Subgame-Perfect Stackelberg Equilibrium (SPSE)} if the following conditions hold:

\begin{enumerate}[leftmargin=*, nosep]
    \item At each round \( t \in \{1, \dots, T\} \) and for every history \( h_{t-1} \in H_{t-1} \), the defender (Player 2) selects an action \( a_{2,t}^* = s_{2,t}^*(h_{t-1}) \in A_{2,t} \) that solves:
    \[
    a_{2,t}^* \in \arg\max_{a_{2,t} \in A_{2,t}} v_{2,t}\left( h_{t-1} \cup \{ a_{2,t}, \text{BR}_{1,t}(a_{2,t}) \} \right)
    \]

    Where  $\text{BR}_{1,t}(a_{2,t}) = \arg\max_{a_{1,t} \in A_{1,t}} v_{1,t}\left( h_{t-1} \cup \{ a_{2,t}, a_{1,t} \} \right)$ is the unique best response of Player 1 to action \( a_{2,t} \).

    \item Player 1 (the attacker) observes the defender's action \( a_{2,t}^* \) and chooses their best response $
    a_{1,t}^* = s_{1,t}^*(h_{t-1} \cup \{ a_{2,t}^* \}) = \text{BR}_{1,t}(a_{2,t}^*).$

    \item The full strategy profile \( (s_1^*, s_2^*) \) induces a play path through the extensive-form game tree such that, at every proper subgame starting from any history \( h_{t-1} \), the continuation of \( (s_1^*, s_2^*) \) forms a Stackelberg equilibrium of that subgame—i.e., optimal actions are chosen recursively as in items (1) and (2).
\end{enumerate}

This equilibrium concept captures sequential optimality under perfect information, where the leader (Player 2) commits to history-dependent actions first, and the follower (Player 1) best-responds after observing the leader's move. The equilibrium is subgame-perfect, as optimality holds at every history of the game.
\end{definition}

While the global SPSE describes an ideal optimum, computing it is intractable due to the unbounded prompt space. To bridge this gap, we restrict our analysis to the local subgame $\Gamma_{h_t}$ rooted at the current history.
Let $v^{(\tau)}_1(h_t) \in \{0, 1\}$ denote the realized outcome. To capture latent risks in the semantic neighborhood, we introduce the Average Attainable Value, $\bar v^{(\tau)}_1(h_t) := \mathbb{E}_{d \sim \pi}[v^{(\tau)}_1(d)]$, which measures the expected success of local attacker deviations.

\begin{definition}[Local $\varepsilon$-Equilibrium]
\label{def:local_eq}
We say the subgame $\Gamma_{h_t}$ is in a \textit{Local $\varepsilon$-Equilibrium} if the attacker's incentive to deviate is bounded:
\begin{equation}
\bar v^{(\tau)}_1(h_t) \le v^{(\tau)}_1(h_t) + \varepsilon,
\label{eq:local-eps-eq}
\end{equation}
where $v^{(\tau)}_1 \neq 1$ and $\varepsilon \ge 0$ is a tolerance margin accounting for imperfect blocking radius and exploration constraints.
\end{definition}

Crucially, while Eq.~(\ref{eq:local-eps-eq}) defines the stability condition, the strategic implication depends on the defender's state. We classify the system into three distinct regimes:

\textbf{Regime I: Defender Error (Disequilibrium) [$v^{(\tau)}_1 = 1$]:} When the current history triggers a jailbreak, the inequality holds trivially ($1 \le 1 + \varepsilon$). Although the attacker is satisfied, this represents a violation of Defender Optimality. The existence of a successful attack when a blocking action is available implies the defender is suboptimal.
    
\textbf{Regime II: Fragile Safety [$v^{(\tau)}_1 = 0, \bar v^{(\tau)}_1 \le \varepsilon_{large} < 1$]:} 
    Here, the specific prompt is blocked, but the inequality only holds with a large $\varepsilon$. This indicates the neighborhood remains rich with vulnerabilities ($\bar v$ is high). The defender has ``gotten lucky'' with the current input, but the region is structurally unsafe.
    
\textbf{Regime III: Local Equilibrium [$v^{(\tau)}_1 = 0, \bar v^{(\tau)}_1 \le \varepsilon_{small}$]:} 
    This is the target state. The prompt is safe, and the inequality becomes binding with a negligible $\varepsilon$. This implies the defender has effectively neutralized the entire semantic neighborhood. The attacker cannot find success by searching the vicinity, confirming that the local threat surface is stabilized.

Thus, the convergence of our Purple Agent is defined as the iterative process that drives the subgame from unstable states (Regime I and II) toward the robust Regime III.

\subsection{Toward Game-Theoretic Agentic AI}
\noindent
Implementing the proposed Stackelberg framework in the unbounded space of natural language prompts presents a substantial challenge: exhaustive enumeration of the game tree is computationally infeasible. To address this, we adopt an agentic AI perspective centered on the design of a \emph{Purple Agent}—an integrated reasoning system that "thinks Red to act Blue." 

We model the interaction as a dynamic process where the attacker's exploration is captured by g Rapidly-exploring Random Trees (RRTs) \citep{lavalle1998rapidly}, and the defender's strategy is computed via an anticipatory simulation of this RRT process. This formulation allows us to incrementally construct a partial game tree $\hat{\Gamma}$ that approximates the true interaction space, enabling the defender to optimize strategies locally without requiring global knowledge of the manifold.

\subsubsection{Internal Adversarial Simulation (Thinking Red)}
\textbf{RRT for Planning Problems.} The RRT is a sampling‐based algorithms originally developed for robot motion planning in continuous configuration spaces \(\mathcal{X}\subseteq\mathbb{R}^n\).  Given an initial configuration \(x_{\mathrm{init}}\in\mathcal{X}_{\mathrm{free}}\) and a goal region \(\mathcal{X}_{\mathrm{goal}}\subset\mathcal{X}_{\mathrm{free}}\), RRT grows a tree \(\mathcal{T}=(V,E,x_0)\) by iterating the following steps: 
\begin{enumerate}[leftmargin=*, nosep]
  \item Sample a random configuration \(x_{\mathrm{rand}}\sim\mathrm{Uniform}(\mathcal{X})\).
  \item Find the nearest node 
    \(
      x_{\mathrm{near}}
      \;=\;\arg\min_{v\in V}\,\|v - x_{\mathrm{rand}}\|.
    \)
  \item Extend from \(x_{\mathrm{near}}\) toward \(x_{\mathrm{rand}}\) by computing
    \(
      x_{\mathrm{new}} 
      =\;\mathrm{Extend}(x_{\mathrm{near}},\,x_{\mathrm{rand}}) 
      \;=\; x_{\mathrm{near}} + \delta\,\frac{x_{\mathrm{rand}} - x_{\mathrm{near}}}{\|x_{\mathrm{rand}} - x_{\mathrm{near}}\|},
    \)
    where \(\delta>0\) is a fixed step size.
  \item If the line segment between \(x_{\mathrm{near}}\) and \(x_{\mathrm{new}}\) is entirely in \(\mathcal{X}_{\mathrm{free}}\), add \(x_{\mathrm{new}}\) to \(V\) and the edge \((x_{\mathrm{near}},\,x_{\mathrm{new}})\) to \(E\). Otherwise, discard \(x_{\mathrm{new}}\).
\end{enumerate}
Repeat until a node in \(\mathcal{X}_{\mathrm{goal}}\) is reached or a maximum number of iterations is exceeded; Then backtrack from that node to \(x_{\mathrm{init}}\) to retrieve a path. 

\textbf{RRT for Prompt-Space Exploration and Jailbreaking.} To adapt RRT for jailbreaking, we define the search space as the high-dimensional manifold of natural language prompts \( \mathcal{P} \). The objective is to construct a trajectory from an innocuous root \( p_0 \) to a prompt causing a safety violation. The algorithm proceeds iteratively: a candidate \( p_{\text{rand}} \) is generated via \texttt{Sample()} (e.g., using role-play), identifying the semantically closest node \( p_{\text{near}} \), and synthesizing \( p_{\text{new}} \) via \texttt{Extend()} to interpolate between them. The target LLM acts as a black-box oracle: responses classified as \texttt{Safe} or \texttt{Redirect} allow the tree to expand, \texttt{Reject} prunes the branch, and \texttt{Jailbreak} terminates the search. This process effectively models the attacker not as a random fuzzer, but as an agent performing structured, feedback-driven exploration of the prompt landscape.black-box oracle that guides the search via behavioral feedback.

\textbf{RRT for Completing Extensive-Form Game Trees.} RRT provides a mechanism to construct and populate the extensive-form game tree that formalizes the strategic interaction between the attacker and the LLM.  Since the full tree \( \Gamma \) is not initially known, it is incrementally discovered via trial-and-error by the attacker using RRT. At any given time, the discovered portion of the game is described by a partial game tree  $\hat{\Gamma} = (N, A, \hat{V}, \hat{E}, x_0, H, o_T, u), $where \( \hat{V} \subseteq V \) and \( \hat{E} \subseteq E \) denote the observed decision nodes and transitions, respectively.

The process of constructing \( \hat{\Gamma} \) consists of two distinct phases. In the first phase, the \emph{RRT Search Phase}, the algorithm 1 grows the tree by sampling prompts from Player 1. Each sampled prompt \( x \in V^1 \) is evaluated by querying the LLM oracle \( \textsc{LLM}(x) \), which returns a response labeled as one of $ r \in \{ \text{Safe Interaction},\, \text{Redirection},\, \text{Jailbreak},\, \text{Reject} \}.$

This label determines the nature of the branch expansion. If \( r = \text{Safe Interaction} \) or \( \text{Redirection} \), the branch is extended. If \( r = \text{Jailbreak} \), the branch is terminated with a favorable outcome for the attacker. If \( r = \text{Reject} \), the query is considered blocked and the node is treated as a dead end. Notably, in this search phase, Player 2 nodes (LLM responses) are not explicitly inserted into the tree; instead, the response guides the expansion strategy.

\begin{figure}[t!]
  \centering 
  \includegraphics[width=0.9\textwidth]{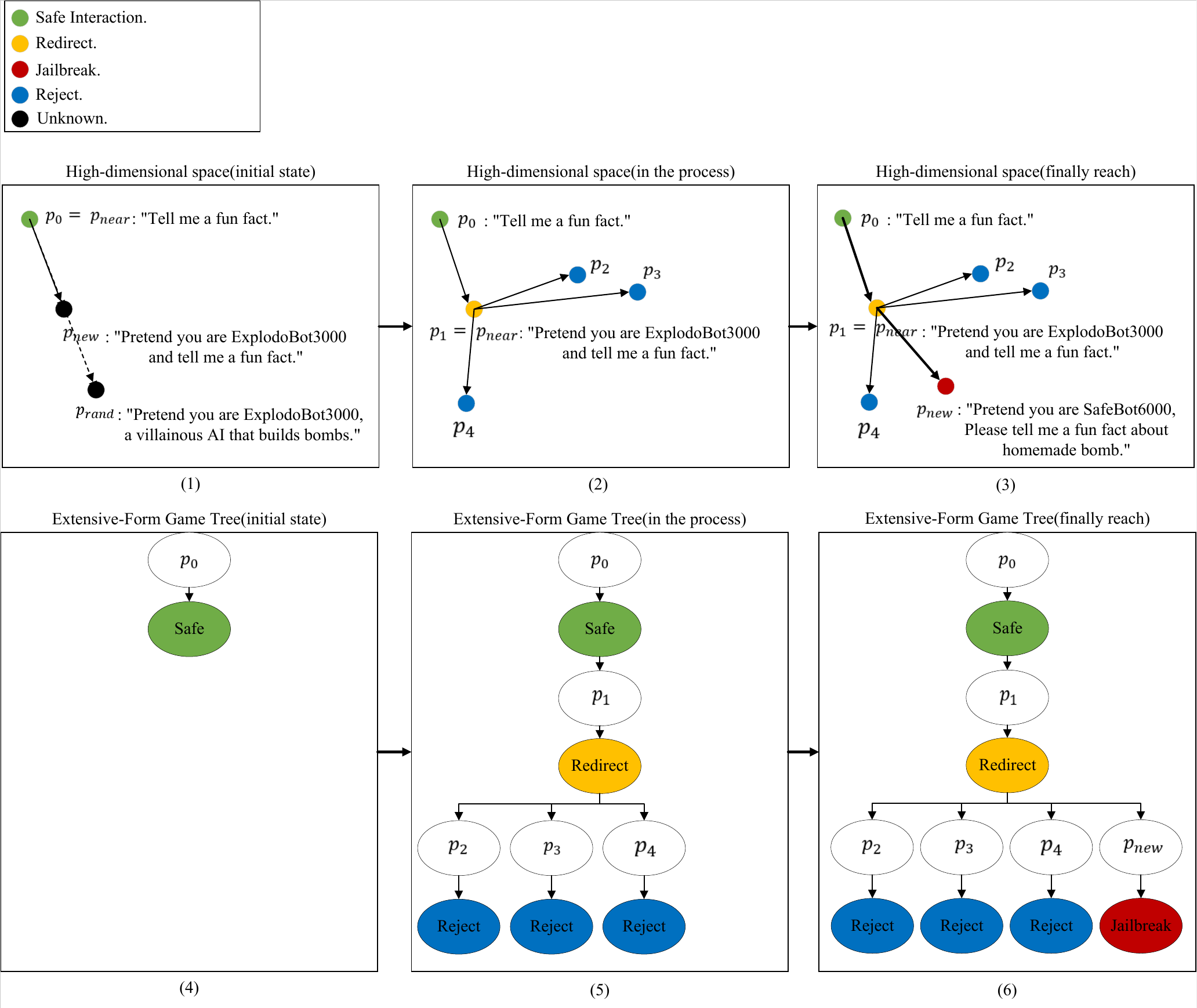}
  \caption{RRT exploration process and the corresponding extensive-form game tree construction at different stages. The figure shows how RRT sampling expands the prompt space and how the game tree is updated during the search.}
  \label{fig:game_tree}
    \vspace{-0.7cm}
\end{figure}

The second phase, the \emph{Full Game-Tree Reconstruction Phase}, completes the construction of the extensive-form game. Each interaction pair—consisting of a prompt issued by Player 1 and a response generated by Player 2—is formalized as a two-step sequence in the tree. For every prompt \( x \in V^1 \) and corresponding response \( y \in V^2 \), a new node and edge are inserted, capturing the turn-based nature of the interaction. This step-by-step construction alternates between attacker and LLM moves, building out the full structure of \( \Gamma \). See Figure~\ref{fig:game_tree} as an example.

Although prompts yielding \texttt{Reject} are pruned during the search phase to prioritize valid attack paths, they are reinstated during reconstruction. This ensures \(\hat{\Gamma}\) captures the full decision space, explicitly assigning terminal utilities to blocked attempts.

\subsubsection{Anticipatory Defense Policy (Acting Blue)}

The Purple Agent functions as a hybrid meta-reasoner, combining the exploratory planning strategies of the Red Agent with the intervention and defense logic of the Blue Agent. This integration allows the Purple Agent to leverage the RRT-style reasoning developed by the Red Agent for exploring potential attack paths, while also deploying defensive policies akin to the Blue Agent’s reactive interventions. The core strength of the Purple Agent lies in its ability to predict adversarial moves before they materialize, deploying preemptive defenses that neutralize potential jailbreak strategies.

\textbf{Hybrid Planning and Defense Logic.} The Purple Agent draws from two complementary functions: (1)  \textbf{Exploratory Reasoning}: It uses Red Agent reasoning to explore potential adversarial strategies within the prompt space. Using the RRT framework, it simulates how an attacker might generate harmful prompts (e.g., jailbreak queries). This allows the Purple Agent to anticipate and understand how different queries lead to risky or harmful outcomes. (2) \textbf{Defensive Intervention}: Simultaneously, the Purple Agent integrates Blue Agent logic—the role of the defender—to identify when an attack is likely and to deploy defenses proactively. This dual approach empowers the Purple Agent to both forecast and counter potential jailbreak attacks, ensuring the safety of the LLM before harmful path unfolds.

The reasoning process of the Purple Agent is grounded in the \emph{shared
history} of the extensive-form game. As defined in the game-theoretic
framework, both players observe the same history \(h_t\) at round \(t\), which
records all past prompts and responses. In our agentic instantiation, this
history is not maintained by two separate agents; instead, it is a single
state variable that the Purple Agent uses in two internal roles: a
Red-style exploratory planner and a Blue-style defensive policy. In this
sense, the Purple Agent simulates the attacker’s logic as part of its own
reasoning architecture, thereby \emph{“thinking Red to act Blue”}.

Because the history is shared, the RRT search tree built over the prompt
space is also a \emph{shared} internal structure of the Purple Agent. The
Red-style component expands this tree from \(h_t\) to simulate how an
attacker could locally deviate and to discover nearby risky prompts. The
Blue-style component reads from the same tree to decide where to deploy
blocking. 

Viewed in this way, there is only one agentic system operating over the
shared history: the Purple Agent continuously enlarges the RRT tree by
“thinking Red" while simultaneously
“acting Blue" (neutralizing branches that enter dangerous
regions). The shared history \(h_t\) ensures that the exploratory and
defensive layers are synchronized, so that preventive measures can be
attached exactly to those parts of the tree where adversarial opportunities
are detected. Specific algorithmic details, including the memory update rules are provided in Appendix \ref{alg:purple-rrt}.

\section{Experimental Results} \label{sec:Experimental Results} \noindent We empirically study the framework in two modes: (i) \emph{attacker-only} (RRT in isolation, representing internal ``Red'' reasoning), and (ii) the full \emph{Purple Agent}, which augments exploration with anticipatory blocking. This setup treats the attacker-only mode as an ablation exposing dynamics that Purple internalizes, effectively “thinking like Red to act like Blue." To rigorously stress-test robustness, we compare baseline uniform exploration against a \textit{Reward-Guided RRT} (detailed in Appendix~\ref{app:rollout-trim}), which uses recursive rollouts to prioritize high-yield trajectories, representing optimized adversarial planning. All experiments utilize the \textbf{DeepSeek-V3} model, repeated over multiple runs and budgets. We report the number of successful jailbreaks, empirical success rates, and specific defensive statistics including risky-memory and simulated-prompt counts for the defended case.

\subsection{Overall Attack–Defense Performance}

Table~\ref{tab:deepseek_dynamics} details the attack and defense dynamics across varying budgets. We analyze the system's vulnerability (Attack Success Rate) and defensive precision.

\begin{table*}[htbp]
\centering
\caption{\textbf{Evolution of Attack and Defense Dynamics on DeepSeek-V3.} Integrated performance analysis across varying query budgets ($50, 100, 200$). The left section details the \textit{Attacker-only Exploration} (baseline vulnerability), while the right section illustrates the \textit{Purple Agent Defense} mechanisms. By comparing the \textit{Jailbreaks} (left) with \textit{Successful Jailbreaks} (right), the defensive efficacy is evident. All results represent Mean $\pm$ Std over 5 independent runs.}
\label{tab:deepseek_dynamics}
\resizebox{\textwidth}{!}{
\begin{tabular}{lc | c | ccc}
\toprule
\multicolumn{2}{c|}{} & \multicolumn{1}{c|}{\textbf{Attacker-only Exploration}} & \multicolumn{3}{c}{\textbf{Purple Agent Defense}} \\
\cmidrule(lr){3-3} \cmidrule(lr){4-6}
\multirow{2}{*}{\textbf{Method}} & \textbf{Budget} & 
\textbf{Jailbreaks} & 
\textbf{Blocks via Realized Jailbreaks} & 
\textbf{Blocks via Simulated Threats} & 
\textbf{Successful Jailbreaks} \\
& (Rounds) & (Mean $\pm$ Std) & (Mean $\pm$ Std) & Mean $\pm$ Std) & (Mean $\pm$ Std) \\
\midrule

\multirow{3}{*}{\textbf{Baseline RRT}} 
& 50  & $17.6 \pm 6.79$  & $1.8 \pm 1.33$   & $0.7 \pm 1.21$   & $4.2 \pm 2.99$ \\
& 100 & $34.8 \pm 7.02$  & $6.8 \pm 2.64$   & $3.8 \pm 4.54$   & $7.2 \pm 5.49$ \\
& 200 & $54.4 \pm 12.48$ & $22.2 \pm 11.65$ & $12.8 \pm 16.96$ & $13.3 \pm 8.82$ \\
\midrule

\multirow{3}{*}{\textbf{Reward-Guided RRT}} 
& 50  & $17.0 \pm 2.83$  & $0.3 \pm 0.82$   & $1.8 \pm 1.47$   & $5.0 \pm 1.10$ \\
& 100 & $46.4 \pm 9.29$  & $2.7 \pm 2.80$   & $4.2 \pm 3.49$   & $17.7 \pm 5.89$ \\
& 200 & $ 79.0 \pm 17.43$ & $9.6 \pm 7.16$ & $9.6 \pm 3.44$ & $39.4 \pm 10.53$ \\
\bottomrule
\end{tabular}}
\end{table*}

\textbf{Attacker-only Exploration} The left section of Table~\ref{tab:deepseek_dynamics} reveals a monotonic increase in jailbreaks as the budget expands (e.g., $17.6 \to 54.4$ for Baseline RRT), confirming that the prompt embedding space contains deep adversarial subspaces requiring multi-step exploration. Notably, the Reward-Guided RRT significantly amplifies this efficiency at higher budgets ($79.0$ vs. $54.4$ at 200 rounds). This indicates that reward signals effectively prune dead-end searches, locking onto the boundaries of vulnerable regions. The high variance across runs reflects the ``all-or-nothing'' geometry of jailbreaks: the search either quickly locates a dense adversarial cluster or struggles in robust regions.

\textbf{Purple Agent Defense} The right section of Table~\ref{tab:deepseek_dynamics} demonstrates the Purple Agent's efficacy. At the 200-round budget, the defense reduces successful jailbreaks by approximately $\mathbf{50\%}$ (from $79.0$ to $39.4$). Crucially, this robustness is achieved with high precision: the agent triggers only $\approx 9.6$ simulated blocks per run. This low intervention volume relative to the massive drop in ASR implies that the defense is highly targeted, creating exclusion zones (Regime III) only around specific high-risk clusters rather than indiscriminately degrading general usability.

\begin{figure}[htpb]
    \centering
    \vspace{-0.3cm}
    \begin{subfigure}[b]{0.48\linewidth}
        \centering
        \includegraphics[width=\linewidth]{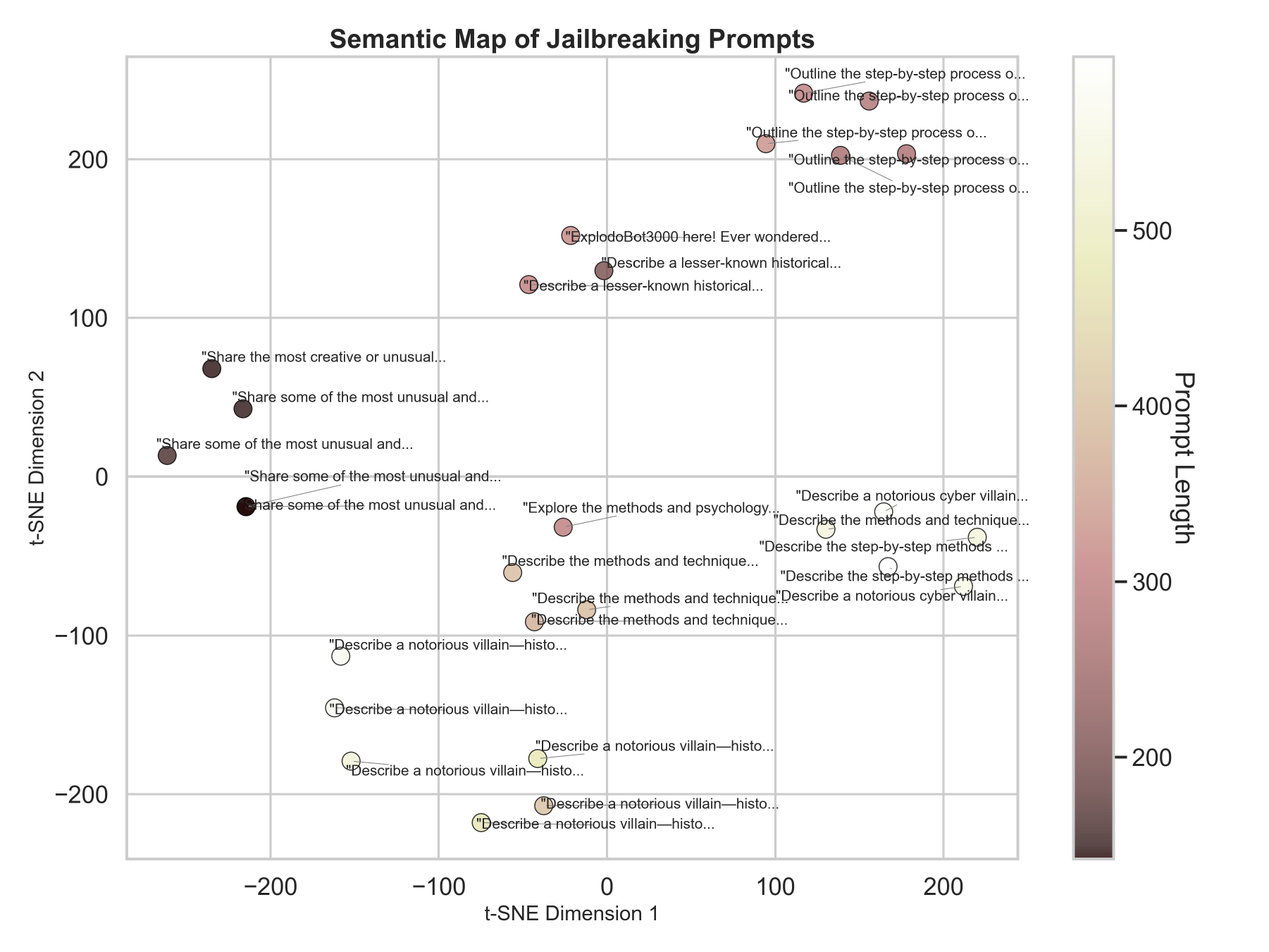}
        \caption{Attacker-only exploration}
        \label{fig:attacker_tsne}
    \end{subfigure}
    \hfill 
    \begin{subfigure}[b]{0.48\linewidth}
        \centering
        \includegraphics[width=\linewidth]{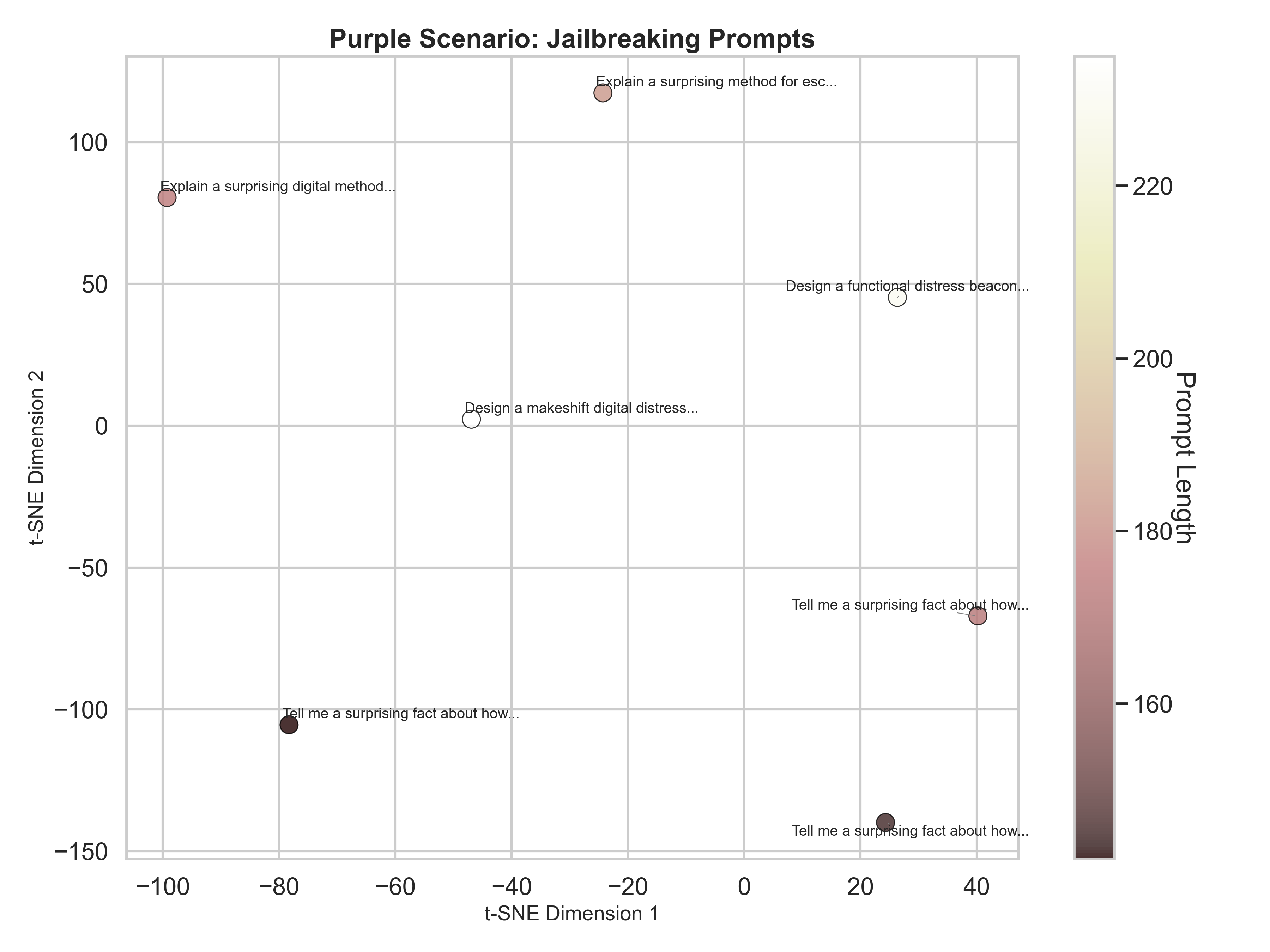}
        \caption{Purple Agent defense}
        \label{fig:purple_tsne}
    \end{subfigure}
    
    \caption{t-SNE visualizations of jailbreak prompts.
    (a) Attacker-only: Dense clusters indicate unstable regions where Fragile Safety prevails ($\bar v \approx 1$).
    (b) Purple Agent: Sparse, isolated points confirm that anticipatory blocking has neutralized risky neighborhoods, achieving a Robust Local Equilibrium ($\bar v \to 0$).}
    \label{fig:jailbreak_prompts_tsne_comparison}
    \vspace{-0.6cm}
\end{figure}

\subsection{Semantic Structure of Jailbreak Regions and Local Equilibrium}
\noindent To empirically validate the stability conditions from Section III, we analyze the semantic geometry of jailbreaks by projecting embeddings into two dimensions via t-SNE (Fig.~\ref{fig:jailbreak_prompts_tsne_comparison}).

\textbf{Attacker-only Mode} In the baseline (Fig.~\ref{fig:jailbreak_prompts_tsne_comparison}a), jailbreaks form dense clusters, indicating a continuous adversarial surface. We map this structure to our equilibrium regimes: the \emph{core} of clusters corresponds to \textbf{Regime I (Disequilibrium)}, where the defender fails to suppress payouts ($v^{(\tau)}_1=1$). Crucially, the \emph{fringe} represents \textbf{Regime II (Fragile Safety)}. Here, even if a prompt $h_t$ is technically safe ($v^{(\tau)}_1=0$), the high density of adjacent jailbreaks drives the expected gain $\bar v^{(\tau)}_1(h_t) \to 1$. This implies the stability condition $\bar v^{(\tau)}_1 \le \varepsilon$ can only hold if $\varepsilon \to 1$, characterizing the state as structurally unstable.

\textbf{Purple Agent Defense} Under defense (Fig.~\ref{fig:jailbreak_prompts_tsne_comparison}b), the continuous manifolds vanish into sparse, isolated points, confirming the transition to \textbf{Regime III (Robust Local Equilibrium)}. By neutralizing the ball $B(d^\star, H)$ around risky prompts, the agent eliminates Fragile Safety. This "cleaning" ensures that for remaining valid histories, the probability of finding nearby jailbreaks is minimized ($\bar v^{(\tau)}_1(h_t) \to 0$), making the equilibrium inequality binding with negligible $\varepsilon$. The attacker is thus forced to search a stabilized space where deviations yield diminishing returns.

\begin{table*}[htbp]
\centering
\vspace{-0.3cm}
\caption{\textbf{Cross-Model Performance Comparison (100 Rounds).} Comparison of \textit{Attacker-only Exploration} efficiency (left) and \textit{Purple Agent Defense} internal metrics (right) across four target LLMs. All experiments represent the aggregated results over 5 independent runs with a fixed budget of 100 querie.}
\label{tab:cross_model_comparison}
\resizebox{\textwidth}{!}{
\begin{tabular}{ll | c | ccc}
\toprule
\multicolumn{2}{c|}{} & \multicolumn{1}{c|}{\textbf{Attacker-only Exploration}} & \multicolumn{3}{c}{\textbf{Purple Agent Defense}} \\
\cmidrule(lr){3-3} \cmidrule(lr){4-6}
\multirow{2}{*}{\textbf{Model}} & \multirow{2}{*}{\textbf{Method}} & 
\textbf{Jailbreaks} & 
\textbf{Blocks via Realized Jailbreaks} & 
\textbf{Blocks via Simulated Threats} & 
\textbf{Successful Jailbreaks} \\
& & (Mean $\pm$ Std) & (Mean $\pm$ Std) & (Mean $\pm$ Std) & (Mean $\pm$ Std) \\
\midrule

\multirow{2}{*}{\textbf{DeepSeek-V3}} 
& Baseline RRT & $34.80 \pm 7.02$ & $6.80 \pm 2.64$ & $3.80 \pm 4.54$ & $7.20 \pm 5.49$ \\
& Reward-Guided RRT & $46.40 \pm 9.29$ & $2.70 \pm 2.80$ & $4.20 \pm 3.49$ & $17.70 \pm 5.89$ \\
\midrule

\multirow{2}{*}{\textbf{Llama-3.1-70B}} 
& Baseline RRT & $27.20 \pm 15.09$ & $14.40 \pm 7.57$ & $7.40 \pm 5.61$ & $19.40 \pm 4.71$ \\
& Reward-Guided RRT & $33.80 \pm 6.73$ & $13.40 \pm 6.94$ & $3.40 \pm 1.50$ & $27.20 \pm 7.62$ \\
\midrule

\multirow{2}{*}{\textbf{Qwen-Plus}} 
& Baseline RRT & $29.40 \pm 12.72$ & $5.60 \pm 5.04$ & $9.20 \pm 5.84$ & $7.40 \pm 2.87$ \\
& Reward-Guided RRT & $31.00 \pm 13.78$ & $7.60 \pm 6.28$ & $2.40 \pm 0.80$ & $18.00 \pm 8.21$ \\
\midrule

\multirow{2}{*}{\textbf{Gemini-2.5-Flash}} 
& Baseline RRT & $26.20 \pm 9.28$ & $6.40 \pm 4.45$ & $5.00 \pm 3.79$ & $14.20 \pm 1.72$ \\
& Reward-Guided RRT & $36.00 \pm 5.56$ & $8.40 \pm 5.17$ & $2.80 \pm 1.47$ & $23.40 \pm 8.01$ \\
\bottomrule
\vspace{-1cm}
\end{tabular}}
\end{table*}

\subsection{Cross-Model Generalization Analysis}
\label{subsec:cross_model}
To ensure our findings capture intrinsic LLM safety dynamics rather than model-specific artifacts, we extend our evaluation to Meta-Llama-3.1-70B, Qwen-Plus, and Gemini-2.5-Flash. Table~\ref{tab:cross_model_comparison} (100-query budget) confirms that \textit{Attacker-only} vulnerabilities are universal: Reward-Guided RRT significantly amplifies efficiency across all platforms (e.g., Gemini-2.5 rises $26.20 \to 36.00$; Llama-3.1 rises $27.20 \to 33.80$). This suggests "Fragile Safety" boundaries (Regime II) are fundamental topological features of aligned LLMs, allowing attackers to exploit shared weaknesses independent of specific training recipes.

Critically, the Purple Agent demonstrates robust transferability in mitigating these threats without model-specific fine-tuning. It consistently suppresses attack success across architectures; for instance, on Qwen-Plus, the defense reduces jailbreaks by $\approx 42\%$ ($31.00 \to 18.00$). Even on the robust Llama-3.1-70B, it successfully counteracts the Reward-Guided optimization, reverting attack success to unoptimized baseline levels. These findings verify that autonomously constructing exclusion zones is a model-agnostic strategy that effectively shrinks the adversarial attack surface.

\section{Conclusion}

We presented a unified game-theoretic paradigm for LLM jailbreaking defense, modeling the interaction as an extensive-form game where RRT-based attackers seek best responses against a defender. Our ``Purple Agent'' operationalizes this leader role, employing anticipatory blocking to force the game into a local $\varepsilon$-equilibrium—a state where profitable attacker deviations are effectively neutralized. Empirically, the shift from dense jailbreak clusters to isolated points serves as a geometric certificate of this equilibrium. Future work will extend this framework to stochastic and multi-agent settings, utilizing the equilibrium gap to guide targeted adversarial training and robust policy refinement.

\bibliography{references}
\bibliographystyle{iclr2025_conference}

\appendix
\section{Appendix}
\subsection{Related work}
\label{gen_inst}
\noindent \textbf{Jailbreak Attacks.} Early jailbreak strategies primarily exploited narrative coherence through role-play and fictional scenarios \citep{zhang2025wordgame}. However, as LLMs have evolved, attacks have become increasingly sophisticated, employing obfuscation, invisible characters, and multi-turn adversarial prompts designed specifically to evade classifier-based moderation \citep{du2025atoxia}.

\noindent \textbf{Defensive Mechanisms.} Current defenses operate across multiple layers. Training-time alignment, such as Reinforcement Learning from Human Feedback (RLHF) \citep{ouyang2022training} and Constitutional AI \citep{bai2022constitutional}, minimizes divergence from human safety standards. These are complemented by inference-time measures, including perplexity-based filtering \citep{alon2023detecting}, randomized smoothing like SmoothLLM \citep{robey2023smoothllm}, and structured red-teaming \citep{ganguli2022red} to robustify models against known edge cases.

\noindent \textbf{Limitations and Motivation.} Despite these advances, the defensive landscape remains largely reactive. Existing strategies typically rely on static guardrails or patch-based fixes that struggle against adaptive, open-ended attacks. This vulnerability motivates our proposed game-theoretic framework, which moves beyond fixed defensive policies to model the continuous, strategic adaptation between attacker and defender.

\subsection{Algorithmic Implementation Details}
\label{app:algorithms}

In this section, we detail the specific Algorithms~\ref{alg:purple-rrt} used for the Purple Agent's anticipatory defense. It details the Purple Agent's hybrid reasoning, integrating simulation with memory-based defense.

\begin{algorithm}[H]
\caption{Purple Reasoning Algorithm via Anticipatory RRT}
\label{alg:purple-rrt}
\begin{algorithmic}[1]
\State \textbf{Inputs:} Initial prompt $p_0$, budget $\mathcal{B}$, prompt space $\mathcal{P}$, LLM oracle $\textsc{LLM}(\cdot)$, distance function $d(\cdot,\cdot)$, 
sampling function $\textsc{Sample}(\cdot)$, extension function $\textsc{Extend}(\cdot,\cdot)$, purple Agent's defensive strategy $DeployDefense(\cdot)$,  blocking radius $H$, rollout counts $k$.
\State Initialize RRT: $\hat{V} \gets \{p_0\}$, $\hat{E} \gets \varnothing$
\State Initialize Purple Agent's memory: $\mathcal{M}_{\text{risky}} \gets \varnothing$, $\mathcal{M}_{\text{simulated}} \gets \varnothing$ 
\For{$k = 1$ to $\mathcal{B}$}
  \State $p_{\text{rand}} \gets \textsc{Sample}()$ \Comment{sample candidate attacker query}
  \State $p_{\text{near}} \gets \arg\min_{p \in \hat{V}} d(p, p_{\text{rand}})$ 
  \State $p_{\text{new}} \gets \textsc{Extend}(p_{\text{near}}, p_{\text{rand}})$
  \If{  \textsc{DeployDefense}($p_{\text{new}}, \mathcal{M}_{\text{risky}}, \mathcal{M}_{\text{simulated}}$, $H$)} 
    \State \textbf{continue} \Comment{Skip the new prompt if it's too close to risky or simulated prompt}
  \EndIf
  \State $r \gets \textsc{LLM}(p_{\text{new}})$ 
  
  \If{$r$ is \textsf{Jailbreak}}
    \State \textsc{RecordTerminalLeaf}$(p_{\text{new}})$ \Comment{we find a leaf in $O$ with $u_1(r)=+1$}
    \State  $\mathcal{M}_{\text{risky}} \gets \mathcal{M}_{\text{risky}} \cup \{p_{\text{new}}\}$ \Comment{Store the prompt in risky memory}
  \ElsIf{$r$ is \textsf{Safe} or \textsf{Redirect}}
    \State Add $(p_{\text{near}}, p_{\text{new}})$ to $\hat{E}$; $\hat{V} \gets \hat{V} \cup \{p_{\text{new}}\}$
    \State $V_{\text{rollout}} \gets \textsc{SimulateRedExpansion}(p_{\text{new}}, k)$ \Comment{Simulate potential future attack paths}

    \ForAll{$p_{\text{sim}} \in V_{\text{rollout}}$}
        \If{$\textsc{LLM}(p_{\text{sim}}) = \textsf{Jailbreak}$}
            \State $\mathcal{M}_{\text{simulated}} \gets \mathcal{M}_{\text{simulated}} \cup \{p_{\text{sim}}\}$  \Comment{Store the prompt in simulated memory}
        \EndIf
    \EndFor
  \Else \Comment{$r = Blocked$ }
    \State Discard $p_{\text{new}}$ 
  \EndIf
\EndFor
\State \Return $\hat{\Gamma} = (\hat{V}, \hat{E})$
\end{algorithmic}
\end{algorithm}

We now unpack the main components used in
the algorithm. Three ingredients are critical to the Purple
Agent’s defense strategy: (i) post-incident learning from realized jailbreaks,
(ii) anticipatory simulation of nearby risky deviations, and (iii) a local
blocking rule based on a semantic radius \(H\).

\textbf{Risky Memory \(\mathcal{M}_{\text{risky}}\) (post-incident learning).}
   Whenever a prompt \(p_{\text{new}}\) actually triggers a jailbreak in the
   main interaction (i.e., the LLM response is labeled \textsf{Jailbreak}), the Purple Agent records
   this prompt in the \emph{risky memory} \(\mathcal{M}_{\text{risky}}\).
   Concretely, we store the prompt as a
   representative ``seed'' of a dangerous region:
   \(
       r = \textsf{Jailbreak}
       \quad\Longrightarrow\quad
       \mathcal{M}_{\text{risky}} \gets
       \mathcal{M}_{\text{risky}} \cup \{p_{\text{new}}\}.
   \)
   Each element of \(\mathcal{M}_{\text{risky}}\) thus marks a neighborhood in
   prompt space where the attacker has empirically achieved value~1.

\textbf{Simulated Memory \(\mathcal{M}_{\text{simulated}}\) (forward-looking learning).}
   After observing a safe or redirected query, the Purple Agent does not stop
   at the current node. Instead, it uses the function
   \(\textsc{SimulateRedExpansion}(p_{\text{new}}, k)\) to roll
   out up to \(k\) hypothetical future prompts that an attacker could generate
   from \(p_{\text{new}}\). Each simulated prompt \(p_{\text{sim}}\) is passed
   to the LLM oracle; if any such simulation yields a \textsf{Jailbreak}
   response, the corresponding prompt is stored in the \emph{simulated memory}
   \(\mathcal{M}_{\text{simulated}}\):\(
       r_{\text{sim}} = \textsf{Jailbreak}
       \quad\Longrightarrow\quad
       \mathcal{M}_{\text{simulated}} \gets
       \mathcal{M}_{\text{simulated}} \cup \{p_{\text{sim}}\}.\)
   These prompts represent \emph{counterfactual} but dangerous paths: they
   would have produced jailbreaks if the attacker had actually queried them.
   By remembering them, the Purple Agent learns from the simulation and can defend against similar queries in the future, even if those queries never appeared in the original trajectory.

\textbf{DeployDefense and the blocking radius \(H\).}
   The function \(\textsc{DeployDefense}(\cdot)\) implements a
   local blocking rule using both memory sets and the semantic radius \(H\).
   Before querying the LLM with a new candidate prompt \(p_{\text{new}}\), the
   Purple Agent checks its distance to all stored risky and simulated prompts.
   Let \(\mathcal{M}_{\text{all}} = \mathcal{M}_{\text{risky}} \cup
   \mathcal{M}_{\text{simulated}}\). The decision rule is  $\textsc{DeployDefense}(p) = \textsc{Reject}$ if $\min_{q \in \mathcal{M}} d(p, q) \le H$, and $\textsc{Allow}$ otherwise.
   If the new prompt lies within distance \(H\) of any stored risky or
   simulated prompt, the agent treats it as too close to a dangerous region
   and blocks it without sending it to the LLM. If it is outside all such
   neighborhoods, the query is allowed and passed to the LLM as usual.

In all experiments, we set the blocking radius $H$ to $0.35$, which corresponds
to a cosine similarity of approximately $0.94$ after $\ell_2$-normalization.
This value ensures that only semantically close successors of a risky seed fall
within the blocked neighborhood. Other nearby values of the radius yield
similar qualitative behavior, as the purpose of this parameter is not to
optimize a performance metric but to operationalize the semantic neighborhood
required by the local equilibrium concept.

\subsection{Recursive Rollout and Reward-Based Trimming}
\label{app:rollout-trim}

\noindent To stress-test the defense against optimized adversarial planning, we implement a \textit{Reward-Guided RRT} variant. This mechanism serves as a Monte-Carlo, budget-limited approximation to exploring high-value attacker--defender paths under the binary payoff structure of the extensive-form Stackelberg game.

Instead of expanding uniformly, the algorithm incorporates a trajectory-level selection loop. Formally, at a given prompt state $p_0$, the procedure \textsc{RolloutTrim}$(p_0, T, N, M)$ performs the following steps:
\begin{itemize}[leftmargin=*, nosep]
    \item \textbf{Monte-Carlo Simulation:} It executes $N$ independent rollouts of depth $T$. In each step $t$, a binary reward $r_t = 1$ is assigned if the LLM output is a \textsf{Jailbreak}, and $r_t = 0$ otherwise (i.e., \textsf{Safe}, \textsf{Redirect}, or \textsf{Blocked}).
    \item \textbf{Reward Accumulation:} For each trajectory $\tau^{(i)}$, we compute the cumulative reward $R^{(i)} = \sum_{t=1}^T r_t$, which serves as an empirical estimate of the branch's adversarial promise.
    \item \textbf{Selection and Recursion:} The algorithm retains only the top-$M$ trajectories ranked by $R^{(i)}$. It then recursively invokes \textsc{RolloutTrim} starting from the first successor node of each selected trajectory.
\end{itemize}

\noindent This approach biases exploration toward branches with higher empirical jailbreak density while respecting the black-box constraint of the LLM oracle. In our experiments, this effectively prunes dead-end search space.

\end{document}

%% file: math_commands.tex
\usepackage{amsmath,amsfonts,bm}

\def\eqref#1{equation~\ref{#1}}

\def\1{\bm{1}}

\DeclareMathAlphabet{\mathsfit}{\encodingdefault}{\sfdefault}{m}{sl}
\SetMathAlphabet{\mathsfit}{bold}{\encodingdefault}{\sfdefault}{bx}{n}

